Xiaowei Xu*, Bi T. Foua, Xingqiao Wang, Vivek Gunasekaran, John R. Talburt

# Leveraging large language models for efficient representation learning for entity resolution

**Abstract:** In this paper, the authors propose TriBERTa, a supervised entity resolution system that utilizes a pre-trained large language model and a triplet loss function to learn representations for entity matching. The system consists of two steps: first, name entity records are fed into a Sentence Bidirectional Encoder Representations from Transformers (SBERT) model to generate vector representations, which are then fine-tuned using contrastive learning based on a triplet loss function. Fine-tuned representations are used as input for entity matching tasks, and the results show that the proposed approach outperforms state-of-the-art representations, including SBERT without fine-tuning and conventional Term Frequency-Inverse Document Frequency (TF-IDF), by a margin of 3–19%. Additionally, the representations generated by TriBERTa demonstrated increased robustness, maintaining consistently higher performance across a range of datasets. The authors also discussed the importance of entity resolution in today's data-driven landscape and the challenges that arise when identifying and reconciling duplicate data across different sources. They also described the ER process, which involves several crucial steps, including blocking, entity matching, and clustering.

**Acknowledgment:** This research was supported in part by the US Census Bureau Cooperative Agreement CB21RMD0160002 for Record Linkage. This project used the facilities provided by the Arkansas High Performance Computing Center supported in part by grants from the National Science Foundation grants #0722625, #0959124, #0963249, #0918970 and a grant from the Arkansas Science and Technology Authority. This project was partially supported by the National Science Foundation under Award No. OIA-1946391.

*Corresponding author: Xiaowei Xu, University of Arkansas, Little Rock, e-mail: xwxu@ualr.edu
**Bi T. Foua, Xingqiao Wang, Vivek Gunasekaran, John R. Talburt,** University of Arkansas, Little Rock

# 1 Introduction

The digital age has ushered in an era in which data are abundant, but the true challenge lies in understanding and processing these data effectively. Representation learning, with its ability to transform raw data into meaningful formats, has emerged as a beacon for this challenge, particularly in fields such as computer vision and information extraction. Automating the discovery of optimal data representations offers a fresh perspective on traditional tasks, thereby enhancing performance and efficiency.

Duplicate data, a pervasive issue in today's landscape, often incur significant costs in terms of money, time, and resources. Different organizations, from businesses to government agencies, grapple with the challenge of identifying and reconciling these duplicates, whether they originate from a single source or multiple disparate sources. This critical task is termed entity resolution (ER), which is a comprehensive process that addresses the challenge of identifying and linking records that refer to the same real-world entity across different data sources. ER becomes particularly complex when data are riddled with inconsistencies or when standardization is lacking. Consider, for example, an e-commerce database in which a single product, such as an "Apple MacBook Pro M2," can manifest in various ways. It may be represented as "Apple, M2 MacBook Pro" in one record or as "MacBook M2 Pro, Apple" in another, despite both referring to the same real-world entity. Similarly, a person named "John Tim Joe, 2022 Sunset Dr apt 217" in one record might appear as "John T Joe, 2022 Sunset Dr" in another record. Such discrepancies require rigorous identification, cleansing, and reconciliation. As described in [1], the ER process involves several crucial steps.

1. *Blocking (or indexing)*: Given the quadratic nature of the ER problem, in which every description should be compared to all others, blocking is applied as an initial step to reduce the number of comparisons. It groups similar descriptions into blocks based on certain criteria, ensuring that comparisons are executed only between descriptions co-occurring in at least one block. This step quickly segments the input-entity collection into blocks, approximating the final ER result.
2. *Entity matching*: This step involves applying a function that determines whether a pair of entity descriptions matches. Typically, a similarity function measures the similarity between two descriptions, with the aim of minimizing false-positive or false-negative matches.
3. *Clustering*: The final task in the ER workflow groups the identified matches together, ensuring that all descriptions within a cluster match. This step infers indirect matching relations among the detected pairs of matching descriptions, thereby overcoming the potential limitations of the employed similarity functions.

A common thread that runs through all these steps of ER, as described above, is the necessity of grouping duplicates together. Whether it is during blocking, entity matching, or clustering, the objective is to bring similar entities closer while pushing dissimilar ones apart. This underlying but important requirement forms the basis of our proposition. In light of this, we unveil TriBERTa, a groundbreaking representation-learning approach tailored for entity resolution. Through TriBERTa, we learn representations that inherently group similar entities together, while separating dissimilar ones, creating a foundational asset that can be leveraged across all steps of the ER process. The learned representation can be applied to each step of the entity resolution process, as shown in Figure 1. This approach harnesses the power of pre-trained language models and representation learning via a triplet loss function. It is important to mention that while our current evaluation focuses solely on entity matching, the potential of TriBERTa extends to other facets of entity resolution, including blocking and clustering, by virtue of its ability to learn representations that group similar entities together and separate dissimilar ones. To encapsulate our contributions:

– We pioneered a representation learning methodology for entity resolution with versatility to span the entire ER process, such as entity matching, data blocking, and data clustering.
– Our research rigorously evaluates the efficacy of learned representations using entity matching as a primary case study within the ER framework.
– A comparative analysis of dedicated end-to-end entity-matching models reveals the robustness and adaptability of our approach across diverse datasets.
– To ensure transparency and reproducibility, we offer extensive data preparation and analysis, enabling peers to replicate and build on our findings.

This chapter is an extension of work originally presented in the paper titled 'Large Language Model-Based Representation Learning for Entity Resolution.

Using Contrastive Learning' [24] which was accepted for presentation at the 2023 International Conference on Computational Science and Computational Intelligence (CSCI) held in Las Vegas, NV, USA.

The remainder of this paper is structured as follows: Section 2 discusses related work with a focus on the entity-matching task, followed by a detailed exposition of our proposed TriBERTa method in Section 3. Section 4 presents the evaluation results, and we conclude with prospective future directions in Section 5.

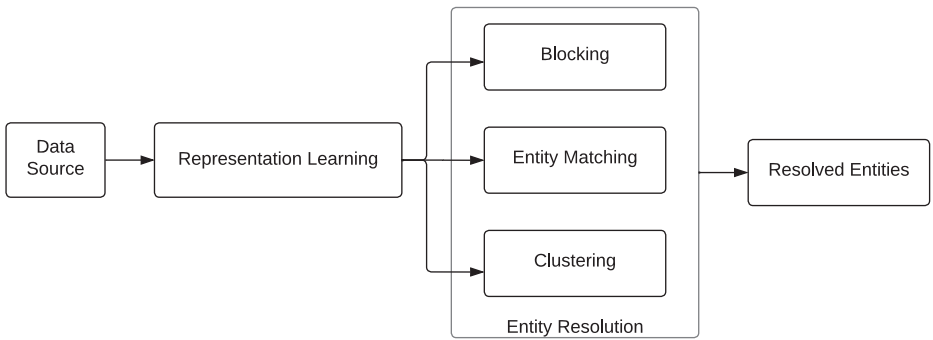

**Figure 1:** Our framework for entity resolution.

## 2 Related work

### 2.1 Entity resolution

Entity resolution has always been a subject of extensive research. While there is no dearth of research on ER, especially in domains like e-commerce, with papers leveraging benchmark datasets such as the "Amazon-Google," "Abt-Buy," "WDC products," "Google-Scholar," "iTunes-Amazon," and ACM datasets [2–6], a significant portion still leans on traditional methodologies.

A glaring gap in the current research landscape is the exploration of representation learning for ERs. Many early studies on ER relied on crowdsourcing approaches. Most crowdsourcing approaches rely heavily on human intervention for proper functioning. Some examples of these crowdsourcing platforms include Amazon Mechanical Turk (AMT) and Crowdflower, which benefited from simple tasks performed by people who were compensated for their efforts. However, crowdsourcing techniques are expensive and unsuitable for production environments [7]. The feasibility of human intervention diminishes as the size of the dataset increases because of the exponential growth in the number of required comparisons.

Subsequent research has helped develop ER-Systems such as Magellan [8, 9] and DeepMatcher [8, 9]. While these systems eliminate the need for human intervention, they often exhibit suboptimal performance (F1-scores), particularly when tested on unseen or noisy data [10]. For instance, on datasets with introduced typos or dropped tokens, Magellan and DeepMatcher yielded unsatisfactory results, rendering them unreliable for use in production.

Recently, the focus has shifted towards more deep learning (DL) approaches for ERs that solely focus on the entity matching task. Notable methods such as KAER[11], JoinBERT[10], SupCon[2], BERT[12], and Ditto[13] have all employed cross-encoders. Although these DL methods have shown promise in achieving good results for entity matching, they often fail to provide embedding for every input record. Having embedding for each record not only facilitates entity matching but also significantly eases the execution of other ER tasks such as clustering or blocking. Therefore, the provision of embedding emerges as a crucial requirement for a holistic and effective ER framework.

Our work with TriBERTa is rooted in the premise of representation learning, which we posit is a pivotal mechanism to bridge the existing gaps in ER tasks. The representations learned through TriBERTa are engineered to group similar entities while separating dissimilar ones, forming a foundational asset that can be leveraged across all steps of the ER process. Although our evaluation in this study is centered on entity matching, the essence of our approach is to demonstrate that a robust representation learning methodology can indeed be a game changer for all facets of ER, including blocking, entity matching, and entity clustering.

## 2.2 Representation learning: contrastive learning

Representation learning has witnessed resurgence in recent years, largely credited to groundbreaking advancements in computer vision and information extraction [14, 15]. In 2018, Wu et al. demonstrated by their experimental results that under unsupervised settings, contrastive representation learning results surpassed the state-of-the-art by a large margin [16]. In 2019, Henaff et al. developed a method that surpassed fully supervised pre-trained ImageNet results using unsupervised contrastive learning to improve transfer learning object detection on the PASCAL VOC dataset [17]. In 2020, Yonglong et al. achieved state-of-the-art results on unsupervised image and video learning benchmarks using contrastive learning [18].

Recently, representation learning, also known as contrastive learning, has been rapidly extended to Natural Language Processing (NLP) for a variety of tasks, including semantic text similarity, semantic search, translation sentence mining, product matching, and notably, entity matching [2, 3]. The fundamental principle of contrastive learning is straightforward: If two records are similar, they are pulled together in the embedding space[1]; conversely, if they are dissimilar, they are pushed apart.

This principle aligns seamlessly with the core objective of entity resolution, which necessitates the grouping of similar entities and separation of dissimilar entities across all its facets – blocking, entity matching, and entity clustering. Contrastive learning, as a DL approach, inherently facilitates the grouping of similar entities in the same embedding space while separating dissimilar entities. This characteristic is pivotal for the effective execution of all ER tasks, underscoring the potential of representation learning and, by extension, contrastive learning as a robust mechanism for advancing ER methodologies. Through this lens, contrastive learning emerges not merely as a technique for enhancing entity matching but as a comprehensive approach capable of significantly improving the broader spectrum of entity resolution tasks.

## 2.3 Triplet loss

Triplet loss is a loss function used in machine algorithms, in which positive and negative inputs are compared to a reference input called an anchor. Triplet loss is a specific type of contrastive learning loss function. The main idea is rooted in the context of nearest neighbor classification [19]. Given a triplet (anchor, positive, negative), the triplet loss function maximizes the difference between the anchor and negative inputs and minimizes the distance between the anchor and positive input. The loss function for one record can be calculated using the Euclidean distance function:

---

**1** embedding space or vectors space is the $k$-dimensional space in which records are represented as vectors.

$$(A, P, N) = \max\left(||f(A) - f(P)||^2 - ||f(A) - f(N)||^2 + \alpha, 0\right)$$

where $A$ is an anchor input, $P$ is a positive input of the same class as $A$, $N$ is the negative input of a different class from $A$, $\alpha$ is the margin (distance) between the positive and negative pairs, and $f$ is a default representation (an embedding).

Figure 2 illustrates the application of triplet loss in FaceNet [19]. In FaceNet, a convolutional neural network (CNN) is trained to optimize the embedding (representations) of the images [20]. A CNN is a type of artificial neural network that is widely used in computer vision and image recognition. As the first two images from the top left represent the same entity, the triplet loss function pulls them together (represented by the inward colliding arrows). For the two pictures in the bottom left, triplet loss increases their difference and pushes them apart (represented by arrows moving apart in opposite directions). This method allows for much greater representational efficiency and better identification of the same images.

The application of triplet loss extends beyond image recognition to the ER domain. By optimizing the representations of entities, triplet loss facilitates the crucial task of grouping similar entities together while separating dissimilar ones, which is a fundamental requirement across all facets of ER – blocking, entity matching, and entity clustering. Learned representations serve as a robust foundation that can be leveraged to enhance the efficiency and accuracy of the ER process.

## 2.4 Sentence BERT

Bidirectional Encoder Representations from Transformers (BERT) language model[2] is an open-source Transformers model[3] for Natural Language processing. It helps the computer understand the meaning of a text or word by using the words or text surrounding it. The BERT language model uses surrounding words or text to establish context. Sentence BERT (SBERT) was first introduced in 2019 by Nils Reimers and Iryma Gurevych as a modification of the pre-trained[4] BERT Language Model [21].

SBERT reduced the computation time used by BERT for finding the most similar pairs of sentences by a factor of 46,800 from 234,000 s (65 h) to 5 s while maintaining the accuracy of BERT [21]. It was mainly developed as a bi-encoder, as opposed to a cross-encoder. Bi-encoders are able to produce embedding for a given sentence. Figure 3 shows a bi-encoder and a cross-encoder. In the bi-encoder, two sentences were independently fed to the BERT. BERT then outputs two vectors (or embedding) $u$ and $v$, which can be compared using a distance metric, such as cosine similarity, Manhattan distance, or Euclidean distance. In contrast, the cross-encoder did not provide any embedding. In the cross-encoder, both sentences are fed to BERT to produce a value between 0 and 1.

---

[2] Language Model is probabilistic distribution over a sequence of words. It helps predict which word is more likely to appear next in a sentence.
[3] Transformers is a novel deep learning architecture that solves sequence-to-sequence NLP tasks and can handle long-range dependencies in texts. Popular language models such as BERT and GPT come from transformers. For a detailed analysis, readers are encouraged to read "Attention is All You Need."
[4] The model is already trained on unlabeled data over similar tasks.

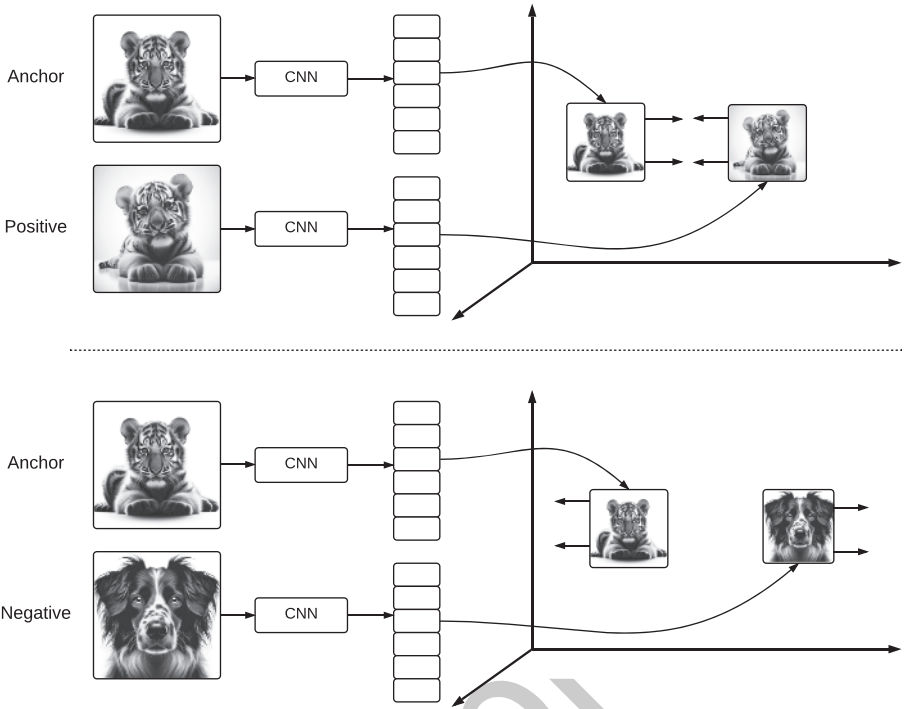

**Figure 2:** Triplet loss representation [19].

SBERT was fine-tuned[5] using Siamese and triplet structure networks to capture meaningful similarities between sentences. The innovation brought about by SBERT in generating embedding for sentences holds significant promise in the domain of entity resolution. By efficiently producing embedding for textual descriptions of entities, SBERT facilitates the grouping of similar entities while separating dissimilar entities across all facets of ER: blocking, entity matching, and entity clustering. The ability to generate meaningful embedding quickly and accurately is a cornerstone of effective representation learning, which, as previously discussed, is pivotal for advancing ER methodologies.

---

**5** the model is first initialized with the pre-trained parameters; then all parameters are fine-tuned using labeled data from the downstream tasks.

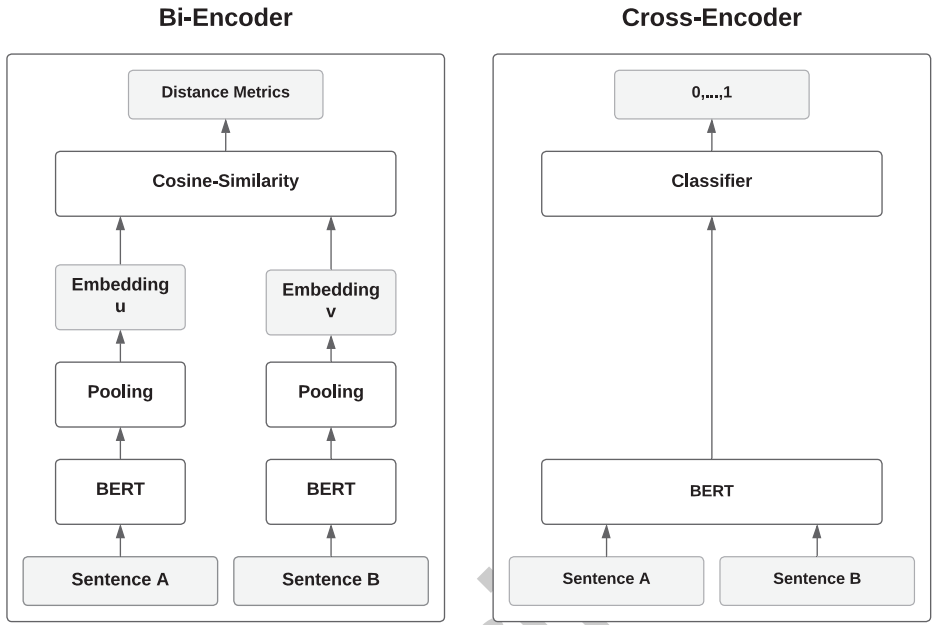

**Figure 3:** Bi-encoder and cross-encoder architecture [22].

## 3 Methodology

As mentioned earlier, although we developed an approach that is applicable to all ER tasks, we only evaluated it on the entity-matching task. Our approach, TriBERTa, comprises two steps:

1. First, the entity records are used as inputs to fine-tune a Language Model and get embedding for every record,
2. Second, a classification task (or entity matching task) is performed using these embedding to determine whether two entities are a match. As our application and evaluation are constrained to entity matching, the second step of this approach is classification. For other facets of the ER, the second step could be blocking or clustering. The embedding architecture we used is similar to the one used in FaceNet; however, instead of images as inputs, we are using text data and replacing the CNN model with a language model, SBERT in our case. In addition, we performed a simple classification task using the embedding obtained from the language model. A simple logistic regression[6] model was used to demonstrate the performance of this approach. Figure 4 shows the overall TriBERTa framework, which is further explained in the following sections.

---

[6] Logistic Regression is a machine learning algorithm used in classification task to analyze the relationship between some dependent variable and a set of independent variables.

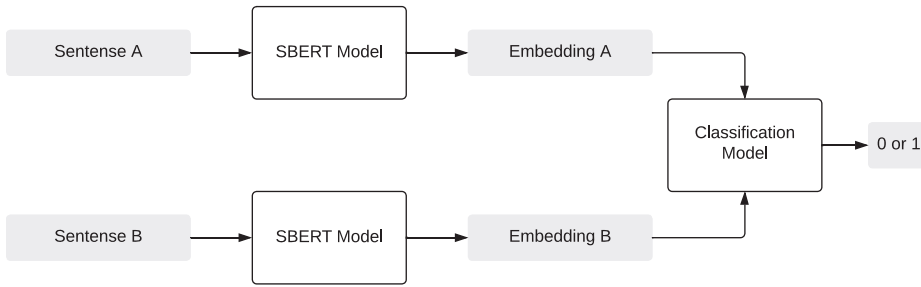

**Figure 4:** TriBERTa framework.

## 3.1 Entity resolution embedding framework using triplet loss

The first step in the overall approach is to determine the embedding (or vectors) for each record. As we know by now, all of the entity resolution steps require that same or duplicate entities be grouped together, so to make sure that every record is mapped correctly in the embedding space we chose triplet loss as a loss function. The goal is to pull together as close as possible vectors that are similar to one another, and push as far as possible vectors that are dissimilar.

As described in Section 2.3, one advantage of contrastive learning through triplet loss is pulling the anchor and positive as close as possible and pushing negative as far as possible from the anchor. Figure 5 shows the framework of the embedding phase. Instead of using a pair of records, an approach used in various studies [2, 3, 12], our embedding framework uses three records or triplets. The three sentences (or records) were fed independently to the SBERT model.

### 3.1.1 Data set preparation for the embedding framework

The embedding framework requires triplet records to be fed into SBERT independently. For this, we modified our datasets to anchor, positive, and negative. Therefore, for each instance (represented as anchor) in the dataset, an instance with the same entity label or id (represented as positive) and another instance with a different entity label (represented as negative) were randomly selected to generate the dataset.

Figure 6 illustrates the data preparation for the triplet loss function. The first table (or origin data) from the left shows that every record has id truth (or Truth ID). The ID truth corresponds to the class of each record. Two records with the same id truths were duplicate records (or the same). As we can observe from the table, the first two records

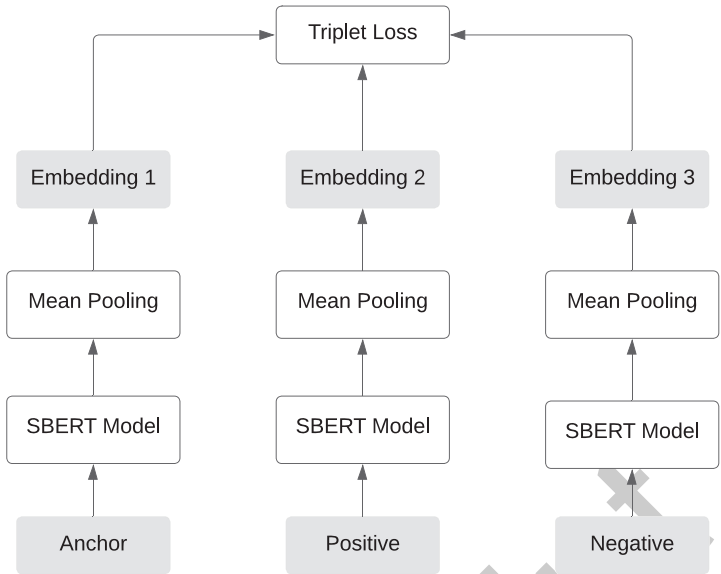

**Figure 5:** Embedding framework.

Jane Mary Doe and Jane M. Doe are duplicates; therefore, they have the same Truth ID. Now that we know duplicate records by their ID truth, we can easily create the second table and third table from the left. For instance, for the first record Jane Mary Doe (our first anchor), we randomly select Jane M. Doe (our first positive) and William P Smith (our first negative). In addition, we dropped rows with null values at the end of the modification.

It is important to note that every positive or negative has an equal probability of being selected. For example, if we were to have another duplicate record for the name Jane Mary Doe, Jane M. Doe and that duplicate would have an equal chance of being randomly selected as a positive.

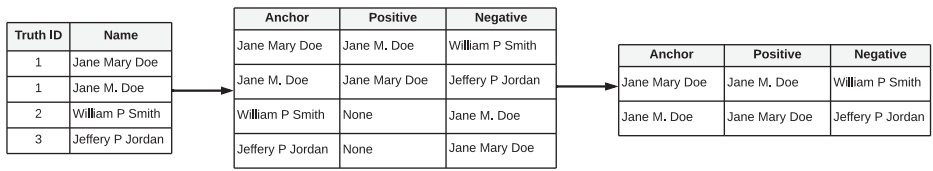

**Figure 6:** Data preparation for triplet loss function.

### 3.1.2 Model selection and fine tuning

Different language models were considered for this task. To select the best model for our embedding task, we chose 14 language models and compared their performance for nearest neighbor search based on cosine similarity. The nearest neighbor should be a positive entity for each anchor.

To choose the best SBERT model, we fine-tuned the 14 language models[7] on a small sample dataset (restaurant dataset of 100 records), similar to our datasets, and compared their nearest neighbor searches for similar batch sizes and epochs. We used the *all-distill RoBERTa-v1,* a RoBERTa model, because it provided the highest accuracy (0.986) of all models tested, as shown in the Appendix table.

To produce a fixed-size output vector for each of our records (or inputs), we added a mean pooling layer (see Figure 5). The mean pooling layer provides the average of all embedding that *all-distill RoBERTa-v1* gives us. This provides us with a fixed embedding vector of 768 dimensions, regardless of the length of the input record.

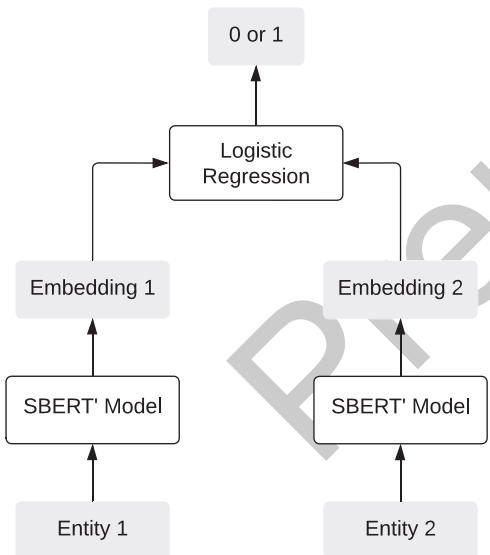

**Figure 7:** Pairwise classification framework.

---

7 Each model has a different parameters and architecture. See appendix table for the list of the models considered and their respective cosine similarity accuracies. More information on each model is available on https://huggingface.co/models or https://www.sbert.net/docs/pretrained_models.html.

## 3.2 Pairwise classification task (entity matching)

The second step in our approach is the application of entity matching, which is a pairwise classification task. As mentioned earlier, although we use entity matching as a second step to evaluate our approach, the second step after obtaining the embedding is blocking, entity matching, or clustering, which are all steps in the entity resolution.

To prove the efficacy of our approach, we used the basic logistic regression model in the second step to classify each pair of entities as a match or no match. Using the fine-tuned *all-distill RoBERTa-v1* model (referred to as SBERT hereafter for simplicity) chosen above from the SBERT package, we find the embedding of each record, which is then fed into the logistic regression model to classify every pair of records as match or no match, as shown in Figure 7.

For this purpose, we modified our datasets for a binary classification task. There are multiple ways to prepare datasets for binary classification tasks. One method is to use the original dataset and select for every record in a dataset a record that matches and another record that does not. Another method uses triplet datasets. For every record composed of anchor, positive, and negative in the triplet, the first two columns (anchor and positive) will represent a match and the first and last columns will represent a no match; therefore, each triplet generates two training samples: one is labeled 1 and the other is labeled 0. We chose the latter (triplet data to the classification dataset) because it maintains a 50/50 split in positive and negative labels, as shown by the following lemma:

$$\forall\, A \in D\; \exists\, P, N \in D$$

that is, for all Anchor A in Dataset D, there exists a positive sentence P and a negative sentence N in dataset D.

This modification is illustrated in Figure 8. Our final classification dataset has twice as many records as the triplet dataset. For the first record, Jane Mary Doe (our first anchor) and Jane M. Doe (our first positive) were selected as positive match with label 1. Jane Mary Doe (our first anchor) and William P Smith (our first negative) are selected as a negative match with label 0.

## 4 Experimental results

In our comprehensive evaluation, we rigorously tested our framework on three widely recognized datasets, benchmarking our results against state-of-the-art representations [21], including the original SBERT model devoid of triplet loss (referred to as RoBERTa) and the conventional TF-IDF method (referred to as TF-IDF). Notably, the underlying model for the SBERT versions remained consistent: the all-distill RoBERTa-v1.

This distinction arose from the fine-tuning process, with our approach leveraging the datasets for this purpose. Both TF-IDF and non-fine-tuned SBERT were employed to derive distinct embedding. These embedding were then processed through one of the simplest yet most effective machine learning models, Logistic Regression, for pairwise classification. Our empirical evaluations underscored the prowess of learned representations in entity matching by setting new benchmarks against established representations.

| Anchor | Positive | Negative |
|---|---|---|
| Jane Mary Doe | Jane M. Doe | William P Smith |
| Jane M. Doe | Jane Mary Doe | William Peters Smith |
| William P Smith | William Peters Smith | Jane M. Doe |
| William Peters Smith | William P Smith | Jane Mary Doe |

| Record 1 | Record 2 | Label |
|---|---|---|
| Jane Mary Doe | Jane M. Doe | 1 |
| Jane Mary Doe | William P Smith | 0 |
| Jane M. Doe | Jane Mary Doe | 1 |
| Jane M. Doe | William Peters Smith | 0 |
| William P Smith | William Peters Smith | 1 |
| William P Smith | Jane M. Doe | 0 |
| William Peters Smith | William P Smith | 1 |
| William Peters Smith | Jane Mary Doe | 0 |

**Figure 8:** Data preparation for the classification task using triplets.

Building on this foundation, we further embarked on a comparative analysis against dedicated end-to-end entity-matching models across various datasets, each presenting unique challenges. This broader evaluation illuminated the robustness and adaptability of our approach, especially when juxtaposed against models tailored specifically for the entity-matching task, reinforcing our belief in the versatility and efficacy of our representation learning approach.

## 4.1 Datasets

The datasets used to test our framework were predominantly sourced from the public domain, but we also explored other datasets, as mentioned in references [11, 13].

Initially, our primary datasets included the GeCo census dataset [23], Cora dataset [23], and restaurant dataset [23]. The GeCo census dataset, which contains address records of people living in the US, was synthetically modified by us to introduce more duplicates. It encompasses 19,993 records with details such as name, address, zip code, city, state, and SSN. The Cora dataset details scientific publications across different topics, with 1,295 records. The restaurant dataset has 868 records, each detailing aspects such as name, address, city, state, zip code, phone number, and other associated data. Table 1 summarizes the initial datasets used in this study. Notably, the GeCo census dataset had the highest duplicate count primarily because of its synthetic nature. This was followed by the Cora dataset and then the restaurant dataset.

To further validate and compare our methods against state-of-the-art cross-encoders dedicated to entity matching only, we employed three additional datasets: GoogleScho-lar, iTunes-Amazon, and ACM, all of which are detailed in references [11, 13]. These datasets, chosen for their unique characteristics and challenges, provided a more comprehensive evaluation platform for our framework. The datasets we used to further test our learning representation framework are structured or "dirty" (or unstructured.) Unstructured data contain noises (missing values, missing characters, etc.) that pose a challenge to any entity-matching model.

**Table 1:** Summary of the first three datasets.

| Dataset | Total number of records | Total number of unique records |
| --- | --- | --- |
| GeCo | 19,993 | 9,996 |
| Cora | 1,295 | 112 |
| Restaurant | 868 | 756 |

## 4.2 Metrics

We considered five metrics to assess the performance of our framework, including cosine similarity for the embedding step, and accuracy, precision, recall, and F1-scores for the classification step.

The metrics used were the common metrics used for classification and embedding tasks. The evaluation metrics are presented in Appendix 6.2 Evaluation Metrics. Cosine similarity is a measure of similarity between two sentences. The closer the value is to 1, the more similar the two sentences are. Accuracy measures the total number of values correctly predicted over the total number of records. Precision is the proportion of actual true positives correctly predicted by the classifier. This is a measure of the true positive rate in the dataset. Recall measures the total number of positive cases from all true positives.

## 4.3 Design of the evaluation

We designed an evaluation to test the performance of both parts of our framework: embedding and classification. We split each dataset into three: training, testing, and validation. For Step 1, the embedding part, we used the validation and training data to fine-tune our SBERT (i.e., *all-distill RoBERTa-v1*) model. For Step 2, the classification task, we used the training and test data to classify pairs of data. Figure 9 illustrates this phenomenon.

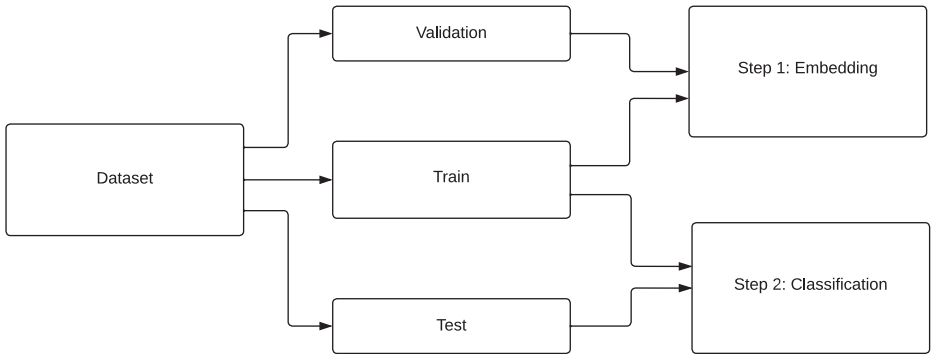

**Figure 9:** Design of evaluation framework.

## 4.4 Results

### 4.4.1 Embedding

Here, we show the results of the first step of our approach for the first three datasets used. In the first step, we trained a pre-trained language model and evaluated it on the validation data. Table 2 shows the training and validation results measured by cosine similarity accuracy, which is a measure of accuracy based on cosine similarity. Using a threshold of 0.5, we achieved an average cosine similarity accuracy of over 99% for all the three datasets. This indicates that the triplet loss methodology improves the representations such that the same entities are recognized as matches and different entities are recognized as non-matches and thus pushed away from one another. This resulted in vectors that were representative of the datasets used in the classification step.

**Table 2:** Triplet loss cosine similarity for the first three datasets.

| Dataset | | Number of records | Cosine similarity (%) |
| --- | --- | --- | --- |
| GeCo | Training | 8,011 | 100 |
| | Validation | 2,670 | 99.9 |
| | Total | 10,681 | – |
| Cora | Training | 765 | 99.0 |
| | Validation | 255 | 98.8 |
| | Total | 1,020 | – |
| Restaurant | Training | 134 | 100 |
| | Validation | 45 | 100 |
| | Total | 175 | – |

**4.4.2 Classification results (entity matching)**

Here, we present the results of the second step of our framework. In the second step, we used the fine-tuned language model TriBERTa as a representation-learning approach and evaluated it on the test data. As depicted in Figures 10–12, the training and test results were measured using accuracy, recall, precision, and F1-score. Performance metrics are captured in Appendix 6.2.

TriBERTa outperformed all baseline methods by a margin of 3–19%. More specifically, TriBERTa improves the F1 measure by 5% when compared with RoBERTa + LR (which is the original SBERT model without fine-tuning for embedding plus logistic regression for entity matching) on average on all three datasets. The largest improvement is achieved in comparison with conventional TF-IDF + LR (which is TF-IDF as embedding plus logistic regression for entity matching), which is over 16% on average for all three datasets. Nevertheless, we observe a slight overfit in the restaurant data, with a drop of almost 5%. We believe that this is because the restaurant dataset was significantly reduced after we modified it to implement TriBERTa. The restaurant had the fewest duplicates, and thus the least amount of data to work with for the classification task. However, despite this deficiency, TriBERTa outperformed TF-IDF and RoBERTa.

Table 3 presents an assessment of the different models across both dirty and structured datasets. A significant metric that captures the eye is the average F1-score, which serves as a holistic indicator of a model's reliability across various datasets. The TriBERTa model recorded an average F1-score of 80.42%. What is remarkable about TriBERTa is not only its good performance on the dirty iTunes-Amazon dataset but also its steadfast consistency. It demonstrated a narrow oscillation in performance, with scores ranging from 72–91.81%. In contrast, the KAER model, which is a cross-encoder technique, attained an average F1-score of 84.82%. While at first glance this may seem commendable, it is crucial to observe the breadth of its performance oscillation. The KAER model exhibited scores that swung from a low of 54.90% to a peak of 98.99%. This wide oscillation suggests a pronounced sensitivity to dataset specifics, raising questions regarding its reliability across diverse real-world datasets. The broad oscillation in scores for models such as KAER might suggest unpredictable behavior when faced with unknown or new datasets.

It is worth noting that the limitation inherent to cross-encoders is their inability to yield embedding. Therefore, they are constrained only to entity matching. On the other hand, consistent performance, as displayed by TriBERTa, is invaluable in practical applications where data variability can challenge models. Another advantage of TriBERTa is its flexibility and wide range of applications. It can be used not only for entity matching, but also for clustering, data blocking, and other NLP tasks.

**Table 3:** F1 score of TriBERTa compared to cross-encoder approaches across different datasets [20].[8]

| Models | Dirty data | | Structured data | | Average |
|---|---|---|---|---|---|
| | GoogleScholar | iTunes-Amazon | iTunes-Amazon | ACM | |
| TriBERTa + LR | 82.90 | 72 | 75 | 91.81 | 80.42 |
| Ditto | 95.54 | 60.38 | 47.06 | 98.55 | 75.38 |
| KAER | 95.74 | 54.90 | 89.66 | 98.99 | 84.82 |

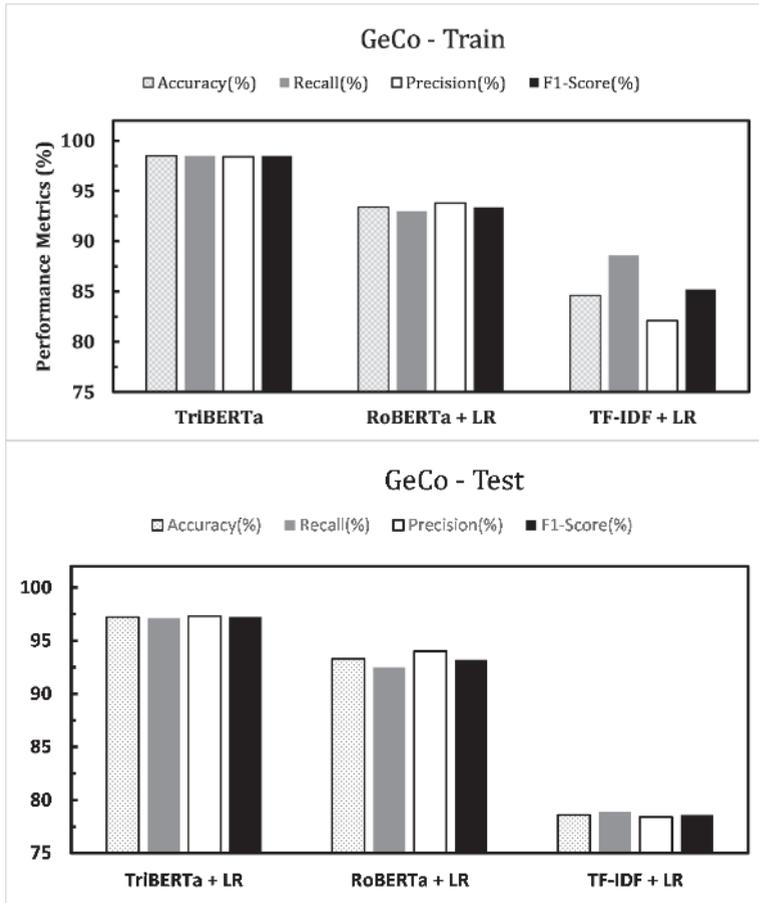

**Figure 10:** GeCo classification model performance.

---

[8] Table data from [20].

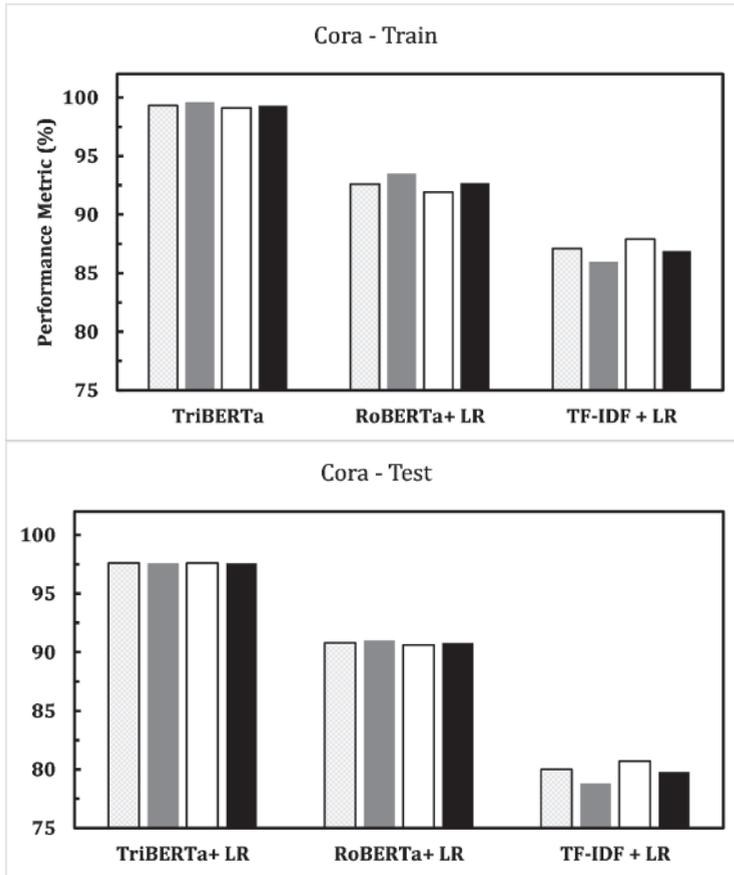

**Figure 11:** Cora classification model performance.

# 5 Conclusion and future work

In the vast landscape of data-driven research, the challenge often lies not in the sheer volume of data but in its effective interpretation and utilization.

Our work with TriBERTa underscores the transformative potential of representation learning to address this challenge, particularly within the realm of entity resolution. Through rigorous empirical evaluations, TriBERTa has demonstrated its capability to not only learn meaningful representations, but also to apply these representations effectively in the context of entity matching.

Our results, which span multiple datasets with varied characteristics, consistently highlight the superiority of TriBERTa's learned representations over traditional and state-of-the-art methods. Furthermore, the robustness exhibited by TriBERTa, especially when compared with dedicated end-to-end entity-matching models, underscores its potential as a versatile tool in real-world scenarios characterized by data heterogeneity.

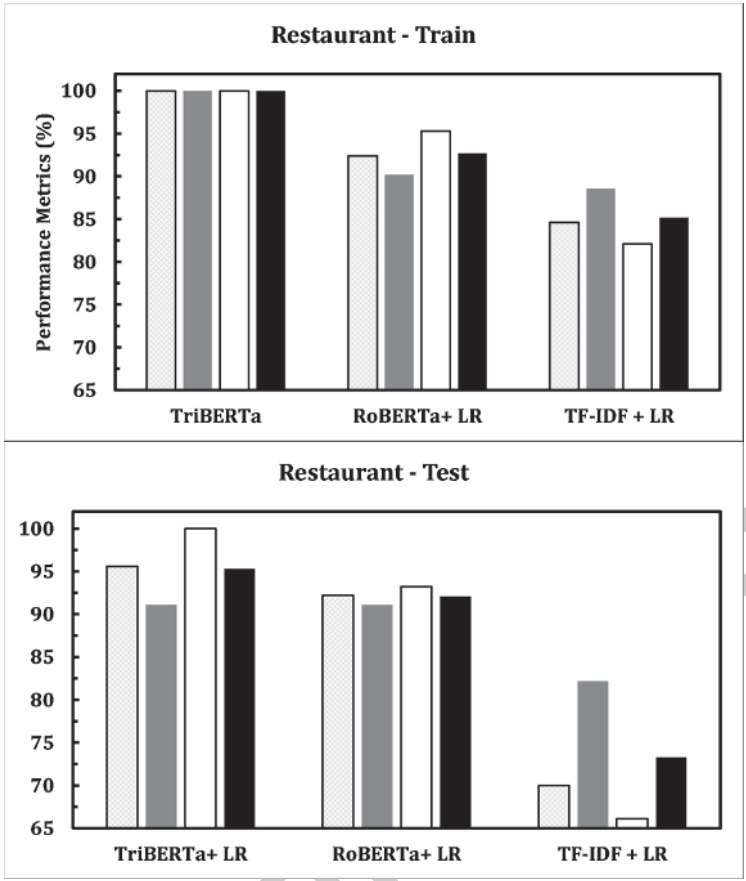

**Figure 12:** Restaurant classification model performance.

While our research is primarily evaluated on entity matching, the foundational principles of representation learning, as embodied by TriBERTa, hold promise for broader applications within the ER process. Future research could explore TriBERTa's efficacy in tasks, such as clustering, data blocking, and other NLP challenges.

In conclusion, TriBERTa stands as a testament to the power of representation learning, offering a fresh perspective on traditional ER tasks and setting the stage for future innovation in this domain.

# 6 Appendix

## 6.1 Models tested

| Model name | Cosine similarity accuracy | Batch size | Epochs |
|---|---|---|---|
| distilBERT-base-uncased | 0.77 | 64 | 10 |
| RoBERTa-base | 0.66 | 64 | 10 |
| all-MiniLM-L12-v2 | 0.94 | 64 | 10 |
| all-MiniLM-L6-v2 | 0.94 | 64 | 10 |
| all-distilRoBERTa-v1 | 0.986 | 64 | 10 |
| all-mpnet-base-v2 | 0.97 | 64 | 10 |
| distiluse-base-multilingual-cased-v1 | 0.93 | 64 | 10 |
| distiluse-base-multilingual-cased-v2 | 0.94 | 64 | 10 |
| multi-qa-MiniLM-L6-cos-v1 | 0.9466 | 64 | 10 |
| multi-qa-distilBERT-cos-v1 | 0.9666 | 64 | 10 |
| multi-qa-mpnet-base-dot-v1 | 0.9666 | 64 | 10 |
| paraphrase-ALBERT-small-v2 | 0.9266 | 64 | 10 |
| paraphrase-multilingual-MiniLM-L12-v2 | 0.90 | 64 | 10 |
| paraphrase-multilingual-mpnet-base-v2 | 0.98 | 32 | 10 |

## 6.2 Entity matching – classification – performance metrics

| GeCo dataset | Model | Accuracy (%) | Recall (%) | Precision (%) | F1-score (%) |
|---|---|---|---|---|---|
| **Train** | TriBERTa | 98.5 | 98.5 | 98.4 | 98.5 |
| | RoBERTa + LR | 93.4 | 93 | 93.8 | 93.4 |
| | TF-IDF + LR | 84.6 | 88.6 | 82.1 | 85.2 |
| **Test** | TriBERTa + LR | 97.2 | 97.1 | 97.3 | 97.2 |
| | RoBERTa + LR | 93.3 | 92.5 | 94 | 93.2 |
| | TF-IDF + LR | 78.6 | 78.9 | 78.4 | 78.6 |

| Cora dataset | Model | Accuracy (%) | Recall (%) | Precision (%) | F1-score (%) |
|---|---|---|---|---|---|
| **Train** | TriBERTa | 99.3 | 99.6 | 99.1 | 99.3 |
| | RoBERTa + LR | 92.6 | 93.5 | 91.9 | 92.7 |
| | TF-IDF + LR | 87.1 | 86 | 87.9 | 86.9 |
| **Test** | TriBERTa + LR | 97.6 | 97.6 | 97.6 | 97.6 |
| | RoBERTa + LR | 90.8 | 91 | 90.6 | 90.8 |
| | TF-IDF + LR | 80 | 78.8 | 80.7 | 79.8 |

| Restaurant | Model | Accuracy (%) | Recall (%) | Precision (%) | F1-score (%) |
|---|---|---|---|---|---|
| **Train** | TriBERTa | 100 | 100 | 100 | 100 |
| | RoBERTa + LR | 92.4 | 90.2 | 95.3 | 92.7 |
| | TF-IDF + LR | 84.6 | 88.6 | 82.1 | 85.2 |
| **Test** | TriBERTa + LR | 95.6 | 91.1 | 100 | 95.3 |
| | RoBERTa + LR | 92.2 | 91.1 | 93.2 | 92.1 |
| | TF-IDF + LR | 70 | 82.2 | 66.1 | 73.3 |

## 6.3 Evaluation metrics

| Metrics | Symbol | Description (or formula) |
|---|---|---|
| True positive | TP | Number of records correctly classified as matches |
| True negative | TN | Number of records correctly classified as non-matches |
| False positive | FP | Number of records incorrectly classified as matches |
| False negative | FN | Number of records incorrectly classified as non-matches |
| Accuracy | Acc | (TP + TN)/(TP + TN + FN + FP) |
| Precision | P | TP/ (TP + FP) |
| Recall | R | TP/ (TP + FN) |
| F1-score | F1 | 2 * (P * R) / (P + R) |
| Cosine similarity | Similarity ($x, y$) | $\cos(\theta) = x.y/||x||*||y||$ |